\documentclass[journal,twoside,web]{IEEEtran}                                                              %

\usepackage{hyperref}

\usepackage[utf8]{inputenc}
\usepackage{graphicx}
\usepackage{algpseudocode}
\usepackage{algorithm}
\usepackage{amsmath,amssymb,amsfonts}
\usepackage{textcomp}
\usepackage{makecell}
\usepackage{algorithmicx}
\usepackage{algpseudocode}
\usepackage{caption}
\usepackage{subcaption}
\algdef{SE}[DOWHILE]{Do}{doWhile}{\algorithmicdo}[1]{\algorithmicwhile\ #1}%
\usepackage[backend=bibtex,style=ieee,maxcitenames=1,mincitenames=1,maxbibnames=3,isbn=false,doi=false,url=false,eprint=false]{biblatex}
\begin{document}

\title{Sim2Real Transfer for Reinforcement Learning without Dynamics Randomization}
\author{Manuel Kaspar, 
	    Juan David Munoz Osorio,
	    Juergen Bock$^{*}$
	    \thanks{$^{*}$Manuel Kaspar, Juan D. M. Osorio and Juergen Bock are part of \textit{Corporate Research} \textit{KUKA Deutschland GmbH} Augsburg, Germany. E-Mail: \{manuel.kaspar,
	    	juandavid.munozosorio, juergen.bock\}@kuka.com}
	}

\maketitle
\begin{abstract}
In this work we show how to use the Operational Space Control framework (OSC) under joint and cartesian constraints for reinforcement learning in cartesian space. Our method is therefore able to learn fast and with adjustable degrees of freedom, while we are able to transfer policies without additional dynamics randomizations on a KUKA LBR iiwa peg-in-hole task.  
Before learning in simulation starts, we perform a system identification for aligning the simulation environment as far as possible with the dynamics of a real robot. Adding constraints to the OSC controller allows us to learn in a safe way on the real robot or to learn a flexible, goal conditioned policy that can be easily transferred from simulation to the real robot.\footnote{This work has been supported by the German Federal Ministry of Education and Research (BMBF) in the project TransLearn (01DQ19007B).}
\end{abstract}

\section{Introduction} \label{sec:introduction}
\begin{figure}[!t]
	\begin{center}
		\includegraphics[height=4.5cm]{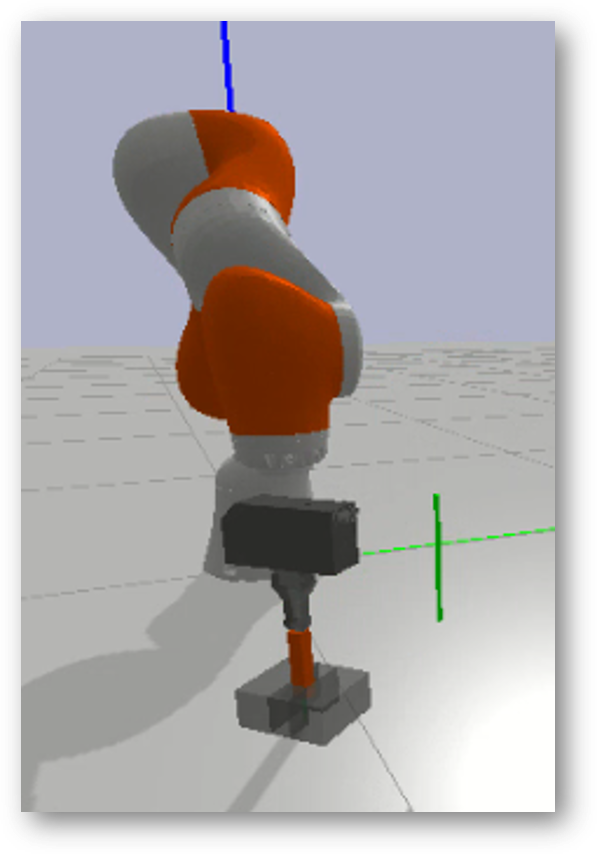}
		\includegraphics[height=4.5cm]{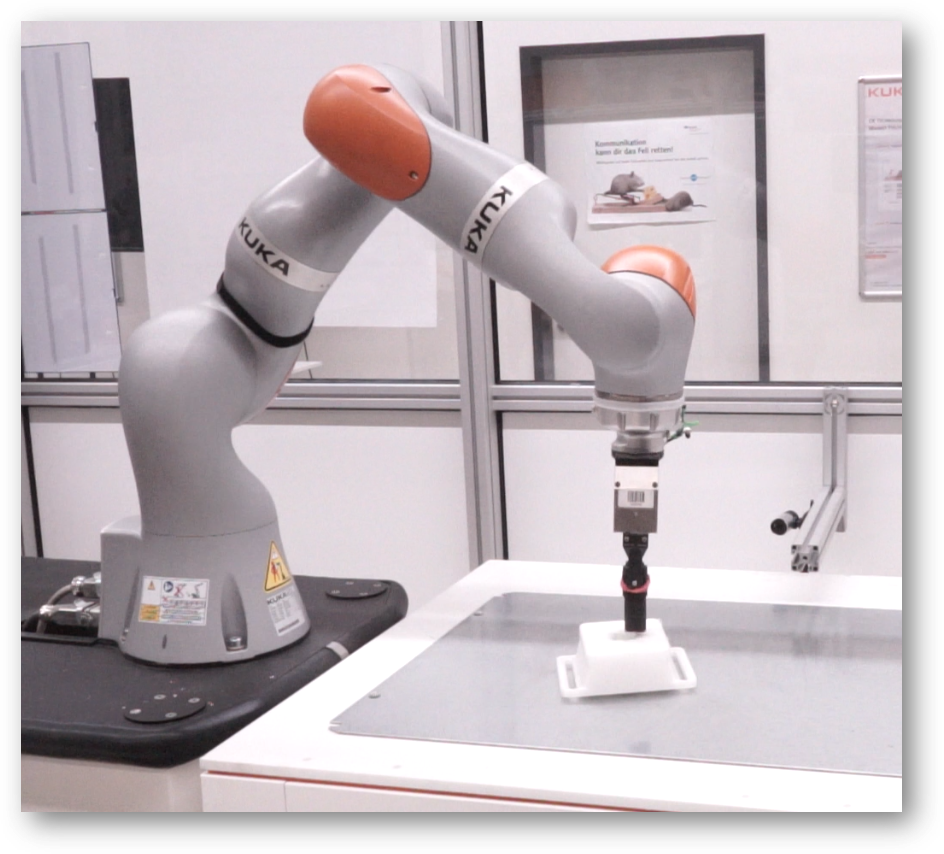}
		\caption{Simulated and real setting}
		\label{fig:realrobot}
	\end{center}
\end{figure}

Most of today's Reinforcement Learning (RL) research with robots is still dealing with artificially simplified tasks, that do not reach the requirements of industrial problems.
This is partly due to the fact that training on real robots is very time-consuming. 
Moreover, it is not trivial to setup a system where the robot can learn a task, but does not damage itself or any task relevant items. 
Therefore, the idea of sim to real transfer~\cite{SimToReal} was introduced.
While this idea seems convincing in the first place, bridging the reality gap is a major difficulty, especially when contact dynamics, soft bodies etc. are involved, where dynamics are difficult to simulate.
This paper investigates possibilities for sim to real transfer while
trying to make the task to learn as easy as possible by using the
Operational Space Control framework (OSC)~\cite{osc}. The controller
takes care of the redundancy resolution and allows to reduce the task dimension. For instance, our current setup tries to perform a peg-in-hole task as shown in Fig.~\ref{fig:realrobot}, where we currently fix two rotational dimensions as we know the required final rotation and just learn the necessary translation and $\theta$-rotation (around the vertical axis) for a successful insertion.

However, pure OSC does not contain information about joint or cartesian limits. We solved that problem by using a novel approach to avoid joint and cartesian limits~\cite{SJSus}~\cite{OSCSCS}. In this way, the output of the controller are joint torques to command the robot that respect joint and cartesian constraints. By limiting not only position but also acceleration and velocity in joint and cartesian space, we avoid damages of the robot or the environment. Moreover, the compliance achieved by using torque control enables the robot to learn tasks, that require robot-environment contacts.

In our opinion those are tasks where RL can bring benefits compared to traditional techniques.
This paper presents a novel approach of integrating RL with OSC, which outperforms traditional approaches that are based on dynamics randomization. Moreover, the combination of RL and OSC bears benefits by avoiding damages of the robot and/or its environment through joint and cartesian constraints.
A video of the results can be found under \url{https://sites.google.com/view/rl-wo-dynamics-randomization}.

\section{Related Work}
Over the past years an increasing number of works tried to use sim to real transfer for learning robotic control:
Progressive Nets~\cite{progressive} were proposed for giving the neural network a flexible way of using or not using past experience which was collected in simulation, when fine tuning on a real system.
Successful sim to real transfer for robots was demonstrated by~\cite{openAiDexterity} and~\cite{OpenAI2019SolvingRC} where in hand manipulation of a cube is learned while also the degree of randomization is adjusted dynamically.
In~\cite{SimToReal} a policy to move an object to a specific position on a table is learned.
The work introduced and analyzed the idea of dynamics randomization in simulation.
Golemo et al.~\cite{simrealnas} try to learn the differences between the real robot and the simulator and then augment the simulation to be closer to the real robot. This is basically a form of system identification, where instead of finding a right set of parameters for a simulator a more sophisticated identification model is learned.
Van Baar et al.~\cite{robustified} perform dynamics randomization for solving a maze game and report easier fine tuning after training a randomized policy in simulation. 
In~\cite{semanticseg} an independent perception and control module is used, while the perception module creates a semantic map of the scene. 
The control module then uses this map as part of its observations. This approach is good for transferring the perception part of a problem from simulation to reality, while the problem of transferring dynamics uncertainties is not discussed in this paper. 
Yan et al.~\cite{rlgrasping2} use Dagger~\cite{dagger} to learn grasping in simulation and by expert demonstration. 
As they perform position control and have a rather easy gripping setup, they do not have to deal with erroneous robot or contact dynamics. 
Like previous work they use a semantic map in their perception module. 
Tan et al.~\cite{simreallocomotion} perform sim to real transfer on learning gates for quadruped robots. 
They use the Bullet~\cite{pybullet} simulation engine (with some improvements) and perform a system identification and dynamics randomization. 
Furthermore, they find that a compact observation space is helpful for sim to real transfer, because the policy can not overfit to unimportant details of the observation. 
Breyer et al.~\cite{rlgrasping} try to learn grasping objects, leveraging an RL formulation of the problem. 
They train on some objects in simulation and then transfer the policy to an ABB YuMI. 
They also use some kind of curriculum learning by starting with a small workspace and then increasing its size.

Inoue et al.~\cite{RlHighPrecision} show how to use a recurrent network to learn search and insertion actions on a high precision assembly task. 
While they achieve success on insertion with high precision requirements, it is only directly applicable to search and insertion tasks. They train two separate networks and need a initial calibration of the system. 
Furthermore, they apply some form of curriculum learning by increasing the initial offset of the peg. 
They do not use a simulation environment but directly train on the robot.
In~\cite{Scherzinger2019Contact} strategies of insertion are learned in task space by using a large number of demonstrations.
We think that our work can figure out strategies more efficiently then leveraging hundreds of demonstrations from humans.
Chebotar et. al~\cite{nvidia} tried estimating parameters of the robot and process from rollouts on the real robot.
In the work of Lee et. al~\cite{lee2019icra} a representation of sensory inputs is learned for performing a peg in hole task, while several sensor modalities are used. 
They use the Operational Space Control framework with an impedance controller and do also command a 4 DOF action vector. 
While using multimodal sensor inputs is an interesting direction, we believe that the insertion performance of our system regarding generalization is comparable to their multimodal system, without additional sensors, while our system runs faster and is more flexible regarding start and target locations.

\section{Reinforcement Learning}
Reinforcement learning is the task to find a policy $\pi(a_t|s_t)$ which selects actions $a_t$ while observing the state of the environment $s_t$. 
The selected actions should maximize a reward $r(s_t, a_t)$. 
The state $s_{t+1}$ and $s_t$ are connected over (stochastic) dynamics $p(s_{t+1}|s_t, a_t)$ which finally creates the trajectory $\tau : (s_0, a_0, s_1, a_1, ..., s_t, a_t)$.

\label{obsvec}
In our case the observation vector $s_t$ contains following variables:

\begin{itemize}
	\item Joint angles $[q_1 ... q_7]$
	\item End effector x, y z positions $[ee_{x}, ee_{y}, ee_{z}]$ 
	\item End effector theta rotation $[ee_\theta]$
	\item End effector velocities $[\dot{ee}_{x}, \dot{ee}_{y}, \dot{ee}_{z}]$
\end{itemize}

The target position of the hole is implicitly encoded into the observation vector. E.g. for the X-dimension $ee_{x} = ee_{x_{cur}} - ee_{x_{target}}$.
$ee_{x_{cur}}$ describes the currently measured X-position of the flange, $ee_{x_{target}}$ the target $x$-position in the hole.
This gives us a goal-conditioned policy.

As an option to give the policy a better hint about the recent history, we also tested stacking $n$ past observations and actions into the observation vector thereby trying to recover the Markov-condition~\cite{rloverview} and giving the network the possibility to figure out the dynamics of the system.\newline

When the observations are stacked we use those values and the last actions and stack it to

\begin{equation}
s = (s_t, a_t, s_{t-1}, a_{t-1}, ..., s_{t-n}, a_{t-n})^T
\end{equation}

\noindent The details of the action vector $a_t$ is described in~\ref{actionvec}.\\

In this work we used the Soft-Actor-Critic (SAC) algorithm explained in~\cite{SAC}. 
We also tried the PPO and DDPG implementation from SurrealAI~\cite{Surreal} but found, that in our experiments SAC was much more sample efficient and stable.

We also investigated the Guided Policy Search algorithm~\cite{Gps} which we found to learn easy tasks really fast. 
Also Levine et al. showed the general applicability to real world robotics tasks and even integrated vision~\cite{Gps2},
we found that the technique strongly depends on the right set of hyperparameters and often fails, when moving to higher dimensional action spaces. 

What makes the Soft-Actor-Critic algorithm so powerful is the fact, that not only a reward $r$ is maximized, but also the entropy of the actor. 
The usage of this maximum entropy framework leads to robust policies, that do not collapse into a single successful trajectory but explore the complete range of successful trajectories. 
This makes the algorithm especially suitable for performing fine tuning on the real robot, after training in simulation.
The objective in the maximum entropy framework is

\begin{equation}
\pi = arg \max_\pi \sum_{t} \mathbb{E}_{(s_t, a_t) \sim p_\pi} [r(s_t, a_t) + \alpha \mathcal{H}(\pi(\cdot|s_t))]
\end{equation}

where $\alpha$ is an automatically adjusted temperature parameter that determines the importance of the entropy term. 
For more details of the SAC algorithm please take a look at~\cite{SAC2}.
The algorithm itself works as shown in~\ref{alg:sac}.

\begin{algorithm}
	\caption{Sampling strategy in the Soft-Actor-Critic algorithm~\cite{SAC2}}
	\label{alg:sac}
	\begin{algorithmic}[1]
		\State Initialize policy $\pi$, critic $Q$ and replay buffer $R$
		\For {$i < max\_iterations$}
			\For {$n < environment\_steps$}
				\State $a_t \sim \pi_\theta(a_t|s_t)$
				\State $s_{t+1} \sim p(s_{t+1}|s_t, a_t)$
				\State $R \gets R \cup {(s_t, a_t, r(s_t, a_t), s_{t+1})}$
			\EndFor
			
			\For {each gradient step}
				\State Get batch from $R$
				\State Update $\pi$ and $Q$ like in Haarnoja et. al~\cite{SAC2}
			\EndFor
		\EndFor
	\end{algorithmic}
\end{algorithm}

SAC is furthermore an off-policy algorithm, what makes it more sample efficient than algorithms like PPO, that also showed to be capable of learning complex policies~\cite{openAiDexterity} and also worked for our task (but slower).

\section{Operational Space Control}
Typically, in OSC, the operational point (in our case, the end effector) is modeled to behave as a unit mass spring damper system:

\begin{equation}\label{eq:impdLaw}
 \boldsymbol{f^* = K e - D \dot{X}},
\end{equation}

where $\boldsymbol{f^*}$ is the command vector, $\boldsymbol{\dot{X}}$ is the vector velocity of the end effector and $\boldsymbol{e}$ is the vector error, that is the difference between the current and the desired offset position of the end effector. $\boldsymbol{K}$ and $\boldsymbol{D}$ are diagonal matrices that represent the stiffness and damping of the system.

RL actions are directly applied on the command vector $\boldsymbol{f^*}$ and are then mapped to the joint space to command the robot using the OSC equation:
 
\begin{equation}\label{eq:maintorque}
 \boldsymbol{\tau = J^T (\Lambda f^* ) + N \tau_{any}},
\end{equation} 
 
where $\boldsymbol{\Lambda}$ is the inertia matrix in the operational space, $\boldsymbol{J}$ is the Jacobian that maps the joint space into the cartesian space and $\boldsymbol{\tau}$ is the vector of command torques that can be send to command the robot. The gravity compensation is done by the lowest torque controller level. Note that the Coriolis terms are despised. In practice, due to inaccuracy of the dynamic model, the performance does not increase by the inclusion of these terms. $\boldsymbol{N = I -J^T \bar{J}^T}$  is the null space projector of $\boldsymbol{J}$ and it exists only for redundant cases (the dimension of $\boldsymbol{f^*}$ is smaller than the number of joints of the robot $n$), with the dynamically consistent Jacobian pseudo inverse $\boldsymbol{\bar{J} = M^{-1} J^T \Lambda }$. $\boldsymbol{\tau_{any}}$ is any torque vector that does not produce any accelerations in the space of the main task, typically choosen to reduce the kinetic energy as $\boldsymbol{\tau_{any} = M(-k_{jointDamp} \dot{q})}$ where $\boldsymbol{k_{jointDamp}}$ is a joint damper term.\\
 
To run a policy on the real robot without breaking or stopping the robot while learning, constraints as joint position and velocity limits should be included in the control algorithm. Cartesian limits are also useful to reduce the work space of learning or to protect the robot to damage itself or objects in the environment. 
 
\subsection{Inclusion of unilateral constraints}
The classic approach to avoid joint limits or cartesian obstacles is to implement potential fields in the proximity to the limit. However, this approach requires a proper setting of the parameters to avoid oscillations or to have a smooth behavior in the control law as shown in~\cite{HanSingAndJoint},~\cite{SJSus}. In~\cite{SJSus}, a simple approach that overcomes these problems is presented. The Saturation in Joint Space (SJS), algorithm~\ref{alg:SJS}, works by estimating the joint accelerations produced by the torque computed from e.g. the impedance law in eq.~\ref{eq:maintorque} (or other task or stack of tasks), and then saturating the joint to its limit (in case of possible violation of the limit). The desired force $\boldsymbol{\Lambda f^* }$ is then achieved at best by the remaining joints. The output of the algorithm is the command torque vector $\boldsymbol{\tau_{c}}$ that respect the joint limits. Note that a Jacobian that maps from the space of the saturated joints to the whole joint space is denoted by $\boldsymbol{J_{lim}}$ and it is defined by:
 
\begin{equation}
\boldsymbol{J_{lim}} =
\begin{bmatrix}
 0 & 1 & 0 & 0 & 0 & 0 & 0 & 0
\end{bmatrix}
\end{equation}
if for instance, the second joint is saturated. To have a better understanding of the SJS approach see~\cite{SJSus}.

\begin{algorithm}
	\caption{Saturation in Joint Space (SJS)}
	\label{alg:SJS}
	\begin{algorithmic}[1]
		\State $\boldsymbol{\tau _{lim} = 0}$ [n$\times$1],
		 $\boldsymbol{N _{lim} = I}$ [n$\times$n]],
		 $\boldsymbol{\ddot{q} _{sat} = 0}$ [n$\times$1] 
		  \Do
		\State $\boldsymbol{\tau_{sjs} = \tau _{lim} + N_{lim} \tau}$ \State $\boldsymbol{\ddot{q} = M^{-1}(\tau_{sjs} -g - c)}$ \State
	   $\ddot{Q} _{max} = min(2 \frac{(Q_{max} - q - \dot{q}dt)}{dt^2},\frac{(V _{max} - q) }{dt}, A_{max})$\State
             $\ddot{Q} _{min} = max(2 \frac{(Q_{min} - q -\dot{q}dt)}{dt^2},\frac{(V _{min} - q )}{dt},A _{min})$
             
             \State $\ddot{q}_{sat,i} =  \left\{ \begin{array}{c}  \ddot{Q}_{max,i} \text{ if } \ddot{q}_i > \ddot{Q}_{max,i} \\ \ddot{Q}_{min,i} \text{ if } \ddot{q}_i < \ddot{Q}_{min,i}  \end{array} \right. $
            \State $\boldsymbol{f^*_{lim} = \ddot{q}_{sat}}$
             \State$\boldsymbol{\tau _{lim}} = \boldsymbol{J^T _{lim} (\Lambda _{lim} f^*_{lim})}$
             \State $\boldsymbol{N_{lim} = I - J^T_{lim}\bar{J}^T_{lim}}$
             \doWhile {$\ddot{q}_i > \ddot{Q}_{max,i}$ or  $\ddot{q}_i < \ddot{Q}_{min,i}$}
	
	\end{algorithmic}
\end{algorithm}

To avoid cartesian limits a similar algorithm to~\ref{alg:SJS} is used \cite{OSCSCS}. The only difference is that everything must be defined in the cartesian space. Algorithm~\ref{alg:SCS} shows how the process works. $\boldsymbol{J_{ev}}$ does the mapping between the sub space of the cartesian space that is being limited and the joint space. For instance, if only the cartesian position is being limited $\boldsymbol{J_{ev}}$ is the first three rows of the whole Jacobian. Note that $\boldsymbol{J_{lim}}$ must do the mapping only from the saturated cartesian space to the Joint space, e.g., it is the third row of $\boldsymbol{J_{lim}}$ if the z dimension gets saturated.

\begin{algorithm}	
	\caption{Saturation in Cartesian space (SCS)}
	\label{alg:SCS}
	\begin{algorithmic}[1]
		\State $\boldsymbol{\tau _{lim} = 0}$ [n$\times$1],
		 $\boldsymbol{N _{lim} = I}$ [n$\times$n]],
		 $\boldsymbol{\ddot{x} _{sat} = 0}$ [l$\times$1] 
		  \Do
		\State $\boldsymbol{\tau_{scs} = \tau _{lim} + N_{lim} \tau}$ \State $\boldsymbol{\ddot{q} = M^{-1}(\tau_{scs} -g - c)}$ 
		\State $\boldsymbol{\ddot{x} = J_{ev} \ddot{q} + \dot{J}_{ev} \dot{q}}$ \State
	   $\ddot{X} _{max} = min(2 \frac{(X_{max} - x - \dot{x}dt)}{dt^2},\frac{(V _{max} - x) }{dt}, A_{max})$\State
             $\ddot{Q} _{min} = max(2 \frac{(X_{min} - x -\dot{x}dt)}{dt^2},\frac{(V _{min} - x )}{dt},A _{min})$
             
             \State $\ddot{x}_{sat,i} =  \left\{ \begin{array}{c}  \ddot{X}_{max,i} \text{ if } \ddot{x}_i > \ddot{X}_{max,i} \\ \ddot{X}_{min,i} \text{ if } \ddot{x}_i < \ddot{X}_{min,i}  \end{array} \right. $
            \State $\boldsymbol{f^*_{lim} = \ddot{x}_{sat}}$
             \State$\boldsymbol{\tau _{lim}} = \boldsymbol{J^T _{lim} (\Lambda _{lim} f^*_{lim})}$
             \State $\boldsymbol{N_{lim} = I - J^T_{lim}\bar{J}^T_{lim}}$
             \doWhile {$\ddot{x}_i > \ddot{X}_{max,i}$ or  $\ddot{x}_i < \ddot{X}_{min,i}$}
	
	\end{algorithmic}
\end{algorithm}

The final control law works by giving  the torque vector $\boldsymbol{\tau}$ from Eq.~\ref{eq:maintorque} to Algorithm~\ref{alg:SCS} as input. The output vector  $\boldsymbol{\tau_{scs}}$ is then given as input to Algorithm~\ref{alg:SJS}. The output vector $\boldsymbol{\tau_{sjs}}$ is then the torque vector that commands the robot. The highest priority is given to the joint limits avoidance that must be respected always. The cartesian limits will be respected as good as they do not interfere with joint limits avoidance.  This control law allows now to learn a policy without breaking the robot or objects in the environment.

\label{actionvec}
The action vector $a_t$ of the learning algorithm consists of $[f^*_x, f^*_y, f^*_z, \theta_{des}]$. Translational commands $f^*_x$, $f^*_y$ and $f^*_z$ are given directly to eq. \ref{eq:maintorque}, while the rotational command $f^*_{\theta}$ is computed by $\theta_{des}$ using eq.\ref{eq:impdLaw}. The error $e$ is calculated in this case by quaternion algebra. Taking $\theta_{des}$ instead of $f^*_{\theta}$ in $a_t$ showed slightly better performance.

\section{Learning flexible cartesian commands by using Operational Space Control}
In our approach we use the OSC to control the robot at torque level ($<=$ 5ms loop) and  do learning on top of this layer (e.g. with 50ms). In detail our control scheme (OSC + SJS + SCS) allows us to have:

\begin{itemize}
	\item Joint limit avoidance
	\item Cartesian walls, where the robot experiences an adversarial force and cannot penetrate them
	\item Velocity saturation (prohibits too fast motions)
\end{itemize}

\subsection{System architecture}
\begin{figure*}[!t]
	\centerline{\includegraphics[width=0.8\textwidth]{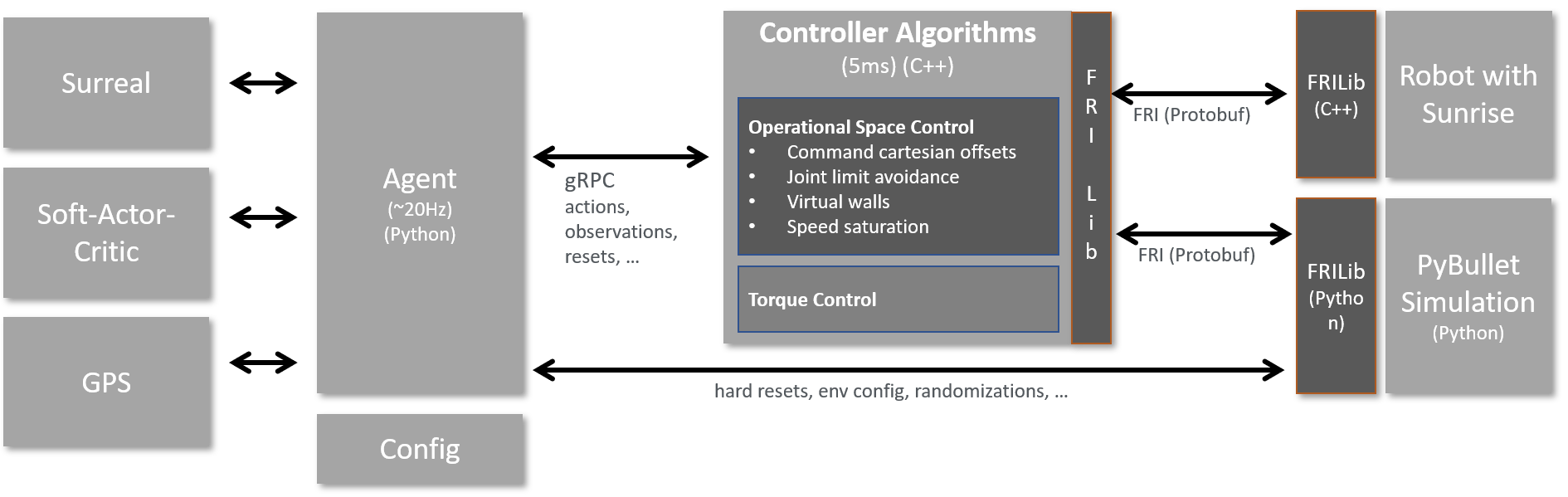}}
	\caption{Architecture for learning and controlling robot and simulation}
	\label{fig:architecture}
\end{figure*}

The system architecture is shown in Fig.~\ref{fig:architecture}. 
We use Python for running reinforcement learning algorithms and PyBullet~\cite{pybullet} for simulation. 
Additionally we have a C++ program that runs the OSC algorithm and uses FRI (KUKA Fast Robotics Interface)~\cite{fri} to command the robot or the simulation.
This enables us to smoothly switch between simulation and the real robot.
The fast C++ implementation ensures that FRI cycle times are met preventing the robot to stop due to timeout errors. 
For the simulation we developed a Python interface for FRI. 
The Python RL algorithm and the C++ controller algorithm communicate over gRPC.

\subsection{Learn task specific cartesian dimensions}
When learning torques it is almost always necessary to learn $n$ joints together to perform an action. The problem increases with complex robots with high number of joints. Nevertheless, tasks like peg-in-hole are almost always easier solvable in cartesian space than in joint space. 
Therefore, we rely on the OSC-framework to map from cartesian commands to torques per joint. 
This gives us a large amount of flexibility to simplify the learning tasks, if necessary.

For instance, if we want to learn a 6 DOF cartesian task, we would still need to learn 7 torque dimensions for the LBR iiwa. 
In cartesian space it is enough to learn the 3 translational dimensions and the 3 rotational dimensions.
If the necessary rotation of a task is clear, this can be given as a fixed setting to the OSC-framework as a task for holding this rotation, and then only the 3 translational dimensions need to be learned. \newline

Therefore every task specific combination is possible:\footnote{$ABC$ is the euler angle notation for rotations, where $A$ rotates around $Z$, $B$ around $Y$ and $C$ around $X$. The frame is expressed in the world coordinate system.}
\begin{itemize}
	\item $XYZ ABC$
	\item $XYZ$ (with fixed rotation)
	\item $XYZ A$
	\item $Z A$
	\item ...
\end{itemize}

$XYZ A$ would, e.g., make sense for a peg-in-hole task where a quadratic object needs to be fitted and a rotation around this axis could be necessary to have the right rotation for aligning peg and hole. A combination $X A$ could, e.g., be used for clipping an electrical component into a rail by performing an approach and rotate/clip motion. 

\section{Sim to Real Transfer}
\subsection{Simulation environment}
We use the PyBullet~\cite{pybullet} simulation environment, where we load an KUKA LBR iiwa 14kg with appropriate dynamics values and an attached Weiss WSG50 gripper. 
We directly command torques to the joints of the robot and use a simulation interval of 5ms. 

\subsection{Dynamics and Environment Randomization}
\cite{openAiDexterity} and~\cite{SimToReal} performed dynamics and environment randomization for being able to transfer their policy from simulation to the real world. 
We found that when using the OSC-framework, system identification and a high-quality model of the robot, we can transfer policies without additional dynamics randomization, which speeds up learning massively and also gives us a higher final performance.
The only parameters we randomize is the start and goal location.

\subsection{System Identification}
In our first trials for using a policy, which was learned in simulation and transferred to the real robot, we found, that it worked pretty poorly. 
The dynamics of the real robot were too different from the dynamics of the simulation. 
Therefore, we performed a special type of system identification, where we run scripted trajectories of actions $a_t$ for $n$ timesteps on the real robot.

Then we used the CMA-ES~\cite{cmaes} algorithm to change the simulation parameters and let them optimize to minimize the 2-norm $(\sum_{i=1}^{n} (v_i)^2)^\frac{1}{2}$ where $v$ is the end effector position.

The optimized simulation parameters are:
\begin{itemize}
	\item Gravity $X$, $Y$, $Z$
	\item Robot link masses scaling
	\item Joint Damping
\end{itemize}

\begin{figure}[!t]
	\centerline{\includegraphics[width=0.8\columnwidth]{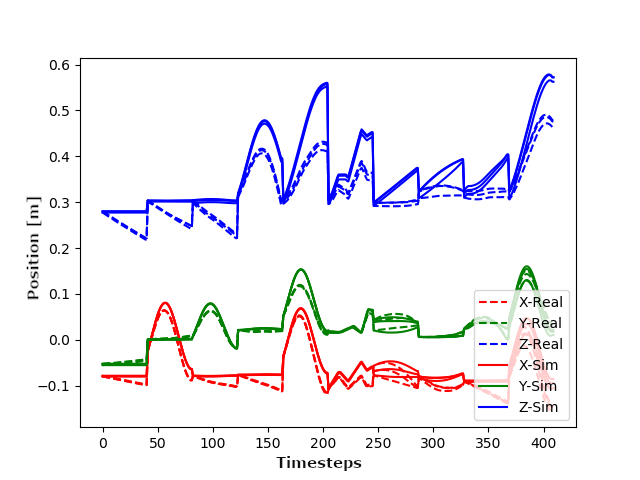}}
	\caption{Real and simulated trajectories at the beginning of the optimization process. Every sub-trajectory consists of 50 time steps (x-axis). The figure shows  10 trajectories behind each other, where 10 different action sequences where chosen. The y-axis corresponds to the position of the flange.}
	\label{fig:identification}
	\bigbreak
	\centerline{\includegraphics[width=0.8\columnwidth]{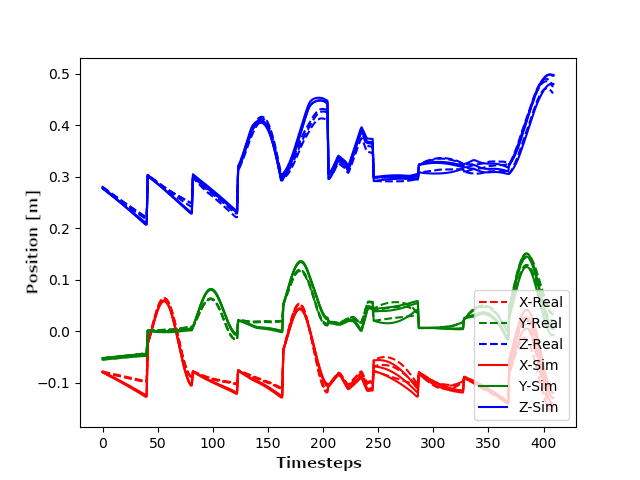}}
	\caption{Real and simulated trajectories after the system identification.}
	\label{fig:identification2}
\end{figure} 

Fig.~\ref{fig:identification} and~\ref{fig:identification2} show the real and simulated trajectory before the system identification and afterwards. We see, that we got much closer to the real trajectory of the robot. 

\section{Evaluation}
In this section we show the results that we found in a simulated environment as well as the results when a policy is transferred to the real robot.
The plots were generated by using five training trials per experiment with a moving average window of 10 and the light-colored background shows the standard deviation of the trials.
In SAC we kept the standard parameters and the maximum number of steps is set to 200, while the episode ends early when the insertion was successful.
We installed and calibrated a camera and an Aruco Marker detector for determining the position and rotation of the hole.
 
By retrieving this position in the control loop and updating the goal
conditioned policy, we achieve to learn a policy that can
interactively react on changes in the goal position during rollouts
and can recover from perturbations (see the video for more details).

As a cost function we used:

\begin{align}
C_{pos} &= \alpha \cdot \| x_{dist}\|_2 + \beta \cdot \| x_{dist}\|_1 + \gamma \cdot \| \theta_{dist}\|_1\\
C_{bonus} &= 50 \text{ if insertion was successful}\\
C_{total} &= - C_{pos} + C_{bonus}
\end{align}

We used $\alpha = 0.6$, $\beta = 0.4$ and $\gamma = 0.1$.

\begin{figure}[!t]
	\begin{center}
		\includegraphics[width=0.9\columnwidth]{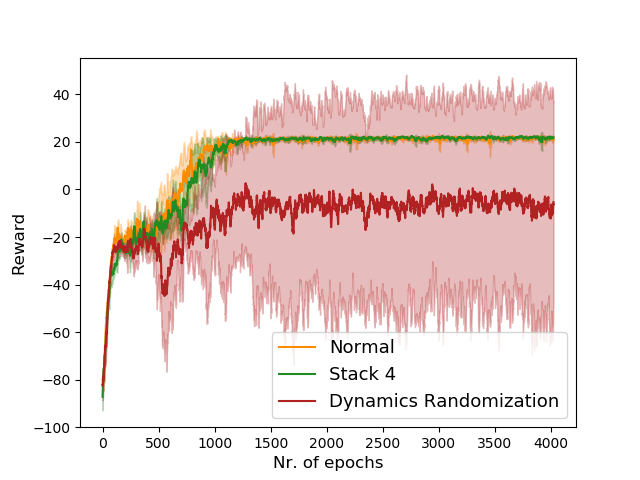}
		\caption{Training with and without dynamics randomization on different start and goal positions. The green plot shows training, when 4 past actions and observations are stacked into the observation vector. The plots show the average and standard deviation over five training runs.}
		\label{fig:training}
	\end{center}
\end{figure}

Training results can be seen in Fig.~\ref{fig:training}. We see that the normal and stacked observation vector perform similarly well in the simulation environment (other training scenarios showed, that this is not always the case and training with stacked observations can slow down and worsen training). 
The red plot shows training, when we perform dynamics randomization. 
Inspired by~\cite{openAiDexterity} we randomize gravity, link masses, joint damping and surface friction. 
We see that the algorithm still mostly succeeds in learning the task but gets much more unstable and sometimes also fails in learning the task at all.

For testing the transfer of the learned policy to the real robot we set the target to three different locations with different $x, y, z, \theta$ the detailed results can be found in Table~\ref{tab:transferresults}.
The unstacked policy transfers slightly better to the real robot and insertion is faster.
We assume this is the case, because overfitting to the simulation could be less serious, when a compact observation space is used like stated in~\cite{simreallocomotion}.
We additionally tried using a different peg-shape (triangle) than the shape for training in simulation.
Insertion with the triangle shape is slightly more difficult.
While insertion with the normal policy works still fine, the performance of the stacked policy degrades.
Transferring the policy which was trained with dynamics randomization does also transfer slightly worse. 

Also training the policy (for one fixed position) directly on the real robot works well (for more details see the video).

These results indicate that a policy trained without dynamics randomization gets trained faster and more reliable and still seems to transfer as well or better than the randomized policy.

\begin{center}
	\captionof{table}{Transfer Results (successful vs. tried insertions)}
	\label{tab:transferresults}
	\begin{tabular}{|c|c|c|c|}
		\hline
		\textbf{Policy} & \textbf{Pos 1} & \textbf{Pos 2} & \textbf{Pos 3} \\
		\hline 
		Stack 0 & $20/20$ & $20/20$ & $20/20$\\ 
		\hline 
		Stack 4 & $20/20$ & $20/20$ & $20/20$\\ 
		\hline 	
		Stack 0 - Triangle & $20/20$ & $20/20$ & $20/20$\\
		\hline
		Stack 4 - Triangle & $18/20$ & $20/20$ & $19/20$\\ 
		\hline 
		Dynamics Randomization & $20/20$ & $20/20$ & $20/20$\\
		\hline 
		Dynamics Rnd. - Triangle & $20/20$ & $20/20$ & $10/20$\\
		\hline
		Train on real robot & $20/20$ & & \\ 
		\hline 
	\end{tabular} 
\end{center}

Additional findings are that policies, which were purely trained in simulation without dynamics randomization are still very robust against perturbations on the real robot. 
For instance, a human can apply forces on the robot arm, while the policy is executed, and it can still recover from those perturbations.
Also moving the target object during execution is possible, as the goal conditioned policy can adapt to the changed situation.
The learned search strategy can find the hole even with perturbations in the target location up to 2\,cm  (if the camera is covered and the hole is moved after the covering). 
The system also learns, that when being below the hole surface it first needs to go over the hole - taking into account preliminary lower reward - to successfully finish insertion.
This is indeed making the problem much more difficult than on plain surfaces and increases training times massively.

\section{Conclusion and future work}
We showed in this work, that it is possible to perform sim to real transfer without doing dynamics randomization. 
This helps speeding up training, can increase performance and reduces the number of hyperparameters. 

In our future roadmap, we plan to investigate the possibilities of using sim to real transfer on more industrial robotic tasks and we believe that our current setup is a good starting point.
In our view, tasks that involve contact are the most interesting class of problems for applying reinforcement learning in robotics. 
They are more difficult to solve, but classic position control tasks can often be solved easier with traditional techniques. 
With today's industrial robots, force sensitive task require a large amount of expert knowledge to program and a big amount of time for fine tuning it to specific applications.
Nevertheless, very often those tasks are also inherently difficult to simulate with today's simulators. 
Friction, soft objects, snap-in events etc. are difficult or even impossible to simulate with tools like PyBullet or MuJoCo. 
Specialized simulation environments that can deal with those challenges in a better way partly exist, but often have other downsides like price or simulation speed. 
We therefore want to investigate how far we can extend sim to real transfer with simulators like PyBullet or MuJoCo on realistic industrial tasks and if industrial requirements for precision, speed and robustness can be met. 

\printbibliography[heading=bibnumbered]

\end{document}